\newif\ifconf\conffalse          

\ifconf
\documentclass[10pt]{IOS-Book-Article}
\usepackage{times,amsmath,amssymb,algorithm,algorithmic}
\normalfont
\usepackage[T1]{fontenc}
\usepackage[dvips]{graphicx}
\else
\documentclass[12pt]{article}
\usepackage{amsmath,amssymb,hyperref,algorithm,algorithmic,graphicx}
\fi

\ifconf
\def\paradot{\subsection}
\else
\newenvironment{keyword}{\centerline{\bf\small
Keywords}\begin{quote}\small}{\par\end{quote}\vskip 1ex}
\def\sep{; }
\def\paradot#1{\vspace{1.3ex plus 0.5ex minus 0.5ex}\noindent{\bf{#1.}}}
\fi

\def\dispmuskip{\thinmuskip= 3mu plus 0mu minus 2mu \medmuskip=  4mu plus 2mu minus 2mu \thickmuskip=5mu plus 5mu minus 2mu}
\def\textmuskip{\thinmuskip= 0mu                    \medmuskip=  1mu plus 1mu minus 1mu \thickmuskip=2mu plus 3mu minus 1mu}
\textmuskip
\ifconf
\def\beq{\vspace{-1ex}\dispmuskip\begin{equation}}    \def\eeq{\vspace{-1ex}\end{equation}\textmuskip}
\def\beqn{\vspace{-1ex}\dispmuskip\begin{displaymath}}\def\eeqn{\vspace{-1ex}\end{displaymath}\textmuskip}
\def\bqa{\vspace{-1ex}\dispmuskip\begin{eqnarray}}    \def\eqa{\vspace{-1ex}\end{eqnarray}\textmuskip}
\def\bqan{\vspace{-1ex}\dispmuskip\begin{eqnarray*}}  \def\eqan{\vspace{-1ex}\end{eqnarray*}\textmuskip}
\else
\def\beq{\dispmuskip\begin{equation}}    \def\eeq{\end{equation}\textmuskip}
\def\beqn{\dispmuskip\begin{displaymath}}\def\eeqn{\end{displaymath}\textmuskip}
\def\bqa{\dispmuskip\begin{eqnarray}}    \def\eqa{\end{eqnarray}\textmuskip}
\def\bqan{\dispmuskip\begin{eqnarray*}}  \def\eqan{\end{eqnarray*}\textmuskip}
\fi

\DeclareMathOperator*{\argmax}{argmax}
\DeclareMathOperator*{\argmin}{argmin}

\begin{document}
\ifconf\begin{frontmatter}\fi

\ifconf
\else

\title{\normalsize\sc Technical Report \hfill IDSIA-05-06
\vskip 2mm\bf\Large\hrule height5pt \vskip 6mm
Metric State Space Reinforcement Learning \\ for a Vision-Capable Mobile Robot
\vskip 6mm \hrule height2pt}
\author{\normalsize {\bf V.\ Zhumatiy} and {\bf F.\ Gomez} and {\bf M.\ Hutter} and {\bf J.\ Schmidhuber}%
\thanks{Also affiliated with: TU Munich, Boltzmannstr.\ 3, 85748 Garching, M\"{u}nchen, Germany.} \\ 
\normalsize IDSIA, \ Galleria 2, \ CH-6928\ Manno-Lugano, \ Switzerland\\
\normalsize \{viktor,tino,marcus,juergen\}@idsia.ch \hspace{11ex} http://www.idsia.ch/}
\date{March 2006}
\maketitle
\fi

\begin{abstract}
We address the problem of autonomously learning controllers for
vision-capable mobile robots. We extend McCallum's (1995)
Nearest-Sequence Memory algorithm to allow for general metrics
over state-action trajectories. We demonstrate the feasibility of
our approach by successfully running our algorithm on a real
mobile robot. The algorithm is novel and unique in that it (a)
explores the environment and learns directly on a mobile robot
without using a hand-made computer model as an intermediate step,
(b) does not require manual discretization of the sensor input
space, (c) works in piecewise continuous perceptual spaces, and
(d) copes with partial observability. Together this allows
learning from much less experience compared to previous methods.
\end{abstract}


\ifconf\end{frontmatter}\else\newpage\fi

\section{Introduction}

The realization of fully autonomous robots will require algorithms
that can learn from direct experience obtained from visual input.
Vision systems provide a rich source of information, but, the
piecewise-continuous (PWC) structure of the perceptual space (e.g.
video images) implied by typical mobile robot environments is not
compatible with most current, on-line reinforcement learning
approaches. These environments are characterized by regions of
smooth continuity separated by discontinuities that represent the
boundaries of physical objects or the sudden appearance or
disappearance of objects in the visual field.

There are two broad approaches that are used to adapt existing
algorithms to real world environments: (1) discretizing the state
space with fixed~\cite{Sutton:98} or
adaptive~\cite{moore94partigame,Pareigis:97} grids, and (2) using
a function approximator such as a
neural-network~\cite{lin:tech92-138,iida:arob03}, radial basis
functions (RBFs)~\cite{anderson:nips5}, CMAC~\cite{sutton:nips8},
or instance-based
memory~\cite{forbes00practical,doya00reinforcement,Santamaria97experiments,smart00practical}.
Fixed discrete grids introduce artificial discontinuities, while
adaptive ones scale exponentially with state space dimensionality.
Neural networks implement relatively smooth global functions that
are not capable of approximating discontinuities, and RBFs and
CMACs, like fixed grid methods, require knowledge of the
appropriate local scale.

Instance-based methods use a {\em neighborhood} of explicitly
stored experiences to generalize to new experiences. These methods
are more suitable for our purposes because they implement local
models that in principle can approximate PWC functions, but
typically fall short because, by using a fixed neighborhood
radius, they assume a uniform sampling density on the state space.
A fixed radius prevents the approximator from clearly identifying
discontinuities because points on both sides of the discontinuity
can be averaged together, thereby blurring its location.  If
instead we use a fixed number $k$ of neighbors (in effect using a
variable radius) the approximator has arbitrary resolution near
important state space boundaries where it is most needed to
accurately model the local dynamics.  To use such an approach,  an
appropriate metric is needed to determine which stored instances
provide the most relevant information for deciding what to do in a
given situation~\cite{kakade03exploration}.

Apart from the PWC structure of the perceptual space, a robot
learning algorithm must also cope with the fact that instantaneous
sensory readings alone rarely provide sufficient information for
the robot to determine where it is (localization problem) and what
action it is best to take.  Some form of short-term memory is
needed to integrate successive inputs and identify the underlying
environment states that are otherwise only {\em partially
observable}.

In this paper, we present an algorithm called Piecewise Continuous
Nearest Sequence Memory (PC-NSM) that extends McCallum's
instance-based algorithm for discrete, partially observable state
spaces, Nearest Sequence Memory
(NSM;~\cite{McCallum:1995instancebased}), to the more general PWC
case. Like NSM, PC-NSM stores all the data it collects from the
environment, but uses a continuous metric on the history that
allows it to be used in real robot environments without prior
discretization of the perceptual space.

An important priority in this work is minimizing the amount of
{\it a priori} knowledge about the structure of the environment
that is available to the learner. Typically, artificial learning
is conducted in simulation, and then the resulting policy is
transfered to the real robot. Building an accurate model of a real
environment is human-resource intensive and only really achievable
when simple  sensors are used (unlike full-scale vision), while
overly simplified models make policy transfer
difficult~\cite{minato}. For this reason, we stipulate that the
robot must learn directly from the real world. Furthermore, since
gathering data in the real world is costly, the algorithm should
be capable of efficient autonomous exploration in the robot
perceptual state space without knowing the amount of exploration
required in different parts of the state space (as is normally the
case in even the most advanced approaches to exploration in
discrete~\cite{Brafman:2003rmax,Kearns:2002poly}, and even in
metric~\cite{kakade03exploration} state spaces).

The next section introduces PC-NSM, section~\ref{sec:exp} presents our
experiments in robot navigation, and section~\ref{sec:discussion}
discusses our results and future directions for our research.

\section{Piecewise-Continuous Nearest Sequence Memory (PC-NSM)}\label{sec:PC-NSM}

In presenting our algorithm, we first briefly review the underlying
learning mechanism, $Q$-learning, then describe Nearest
Sequence Memory which extends $Q$-learning to discrete POMDPs, and
forms the basis of our PC-NSM.

\paradot{Q-learning}
The basic idea of $Q$-learning, originally formulated for finite
discrete state spaces, is to incrementally estimate the value of
state-action pairs, $Q$-values, based on the reward received from
the environment and the agent's previous $Q$-value estimates. The
update rule for $Q$-values is
\beqn
  Q_{t+1}(s_t, a_t) = (1-\alpha) Q_{t}(s_t, a_t) + \alpha [r_{t+1} +
  \gamma \max_a Q_t(s_{t+1},a)]
\eeqn
where $Q_{t}(s_t, a_t)$ is the $Q$-value estimate at time $t$ of
the state $s_t\in S$ and action $a_t \in A$, $\alpha$ is a
learning rate, and $\gamma\in [0,1]$ a discount parameter.

$Q$-learning requires that the number of states $s_t$ be finite
and completely observable.  Unfortunately, due to sensory
limitations, robots do not have direct access to complete state
information, but, instead, receive only observations $o_t\in O$,
where $O$ is the set of possible observations.  Typically, $O$ is
much smaller than the set of states $S$ causing {\em perceptual
aliasing} where the robot is unable to behave optimally because
states requiring different actions look the same.

In order to use $Q$-learning and similar methods under these more
general conditions, some mechanism is required to estimate the
underlying environmental state  from the stream of incoming
observations.  The idea of using the history of all observations
to recover the underlying states forms the core of the NSM
algorithm, described next.

\paradot{Nearest Sequence Memory}
NSM tries to overcome perceptual aliasing by maintaining a
chronologically ordered list or history of interactions between the
agent and environment.  The basic idea is to disambiguate the aliased
states by searching through the history to find those previous
experience sequences that most closely match its recent situation.

At each time step $t$ the agent stores an experience triples $(a_t,
o_t, R_t)$ of its current action, observation, and reward by appending
it to history $h_{t-1} = \langle(a_1, o_1, R_1),\ldots,(a_{t-1},\\
o_{t-1}, R_{t-1})\rangle$ of previous experiences, called {\em
observation state}\footnote{We substitute the symbol $h$ for $s$ in
McCallum's original notation to avoid confusion with the accepted
definition of ``state'' as observation sequences do not correspond to
process states.}.

In order to choose an action at time $T$, the agent finds, for each
possible action $a$, the $k$ observation states in the history that
are most similar to the current situation.
McCallum~\cite{McCallum:1995instancebased} defines
similarity by the length of the common history
\beq\label{eq:nsm-metric}
  n(h_{t},h_{t^{\prime}}) = \begin{cases} 0  &  t = 0 \lor t^{\prime} = 0  \lor
                (a_{t}, o_{t}, R_{t}) \neq (a_{t^{\prime}}, o_{t^{\prime}},
                R_{t^{\prime}}) \\
    1+n(h_{t\!-\!1},h_{t^{\prime}\!-\!1}) &
                (a_{t}, o_{t}, R_{t}) = (a_{t^{\prime}}, o_{t^{\prime}},
                R_{t^{\prime}}).  \end{cases}
\eeq
which counts the number of contiguous experience triples in the two
observation states that match exactly, starting at $t$ and $t^{\prime}$
and going back in time.
We rewrite the original $n$ into a functionally equivalent, but
more general form using the distance measure\footnote{Note that this
$\mu$ is not a metric.} $\mu(h_{t}, h_{t^{\prime}}) =
[1+n(h_{t},h_{t^{\prime}})]^{-1}$ to accommodate the metric we introduce in
the next section.

The $k$ observation states $\mu$-nearest to $h_T$ for each possible
action $a$ at time $T$ form a neighborhood $N_a^{h_T}$ that is
used to compute the $Q$-value for the corresponding action by:
\beq\label{eq:qval}
Q(h_T,{a}) = \left(\frac1{|N_{a}^{h_T}|} \sum_{h_t\in N_{a}^{h_T}}
q(h_t)\right),
\eeq
where $q(h_t)$ is a local estimate of $Q(h_t,a_t)$ at the state-action
pair that occurred at time $t$.

After an action has been selected according to $Q$-values (e.g. the
action $\widehat{a}_T$ with the highest value), the $q$-values are updated:
\beq\label{eq:update}
  q(h_i) :=  (1-\beta)q(h_i) + \beta(R_i + \gamma\max_a Q(h_T,a)),\
  \ \forall h_i \in N_{\widehat{a}_T}^{h_T}.
\eeq
NSM has been demonstrated in simulation, but has never been run on
real robots.  Using history to resolve perceptual aliasing still
requires considerable human programming effort to produce reasonable
discretization for real-world sensors.  In the following we avoid the
issue of discretization by selecting an appropriate metric in the
continuous observation space.

\renewcommand{\algorithmiccomment}[3]{\hspace{#3}
            \parbox[t]{#1}{{\em// #2}}
        }
\begin{algorithm}[t] 
\caption{Piecewise-Continuous NSM}
\label{alg:tra}
\begin{algorithmic}[1]
\STATE $T=0$
\STATE $h_T\Leftarrow\emptyset$
\COMMENT{3in}{initialize history}{0.2in}
\LOOP
  \STATE $T\Leftarrow T+1$
  \STATE $h_T\Leftarrow h_{T-1} + (a_T,o_T,R_T)$
  \COMMENT {3in}{get and store action, observation, and reward}{0.2in}
  \FOR{each $a\in A$}
    \STATE $\mathcal{H}^a = \{h_t | a_{t} = a, \forall t < T\}$
    \COMMENT{2.7in}{split the history into subsets $\mathcal{H}^a$
    containing all \\// observation states where action $a$ was taken}{0.1in}
    \STATE $N_a^{h_T} = \underset{h_t\in \mathcal{H}^a}\argmin^k
        \mu (h_t, h_T)$
      \COMMENT{2.5in}{find the $k$-nearest neighbors to $h_T$ in
      $\mathcal{H}^a$ \\// using metric $\mu$ from
      equation~\ref{eq:discounted-metric}}{0.2in}

    \STATE $Q(h_T,{a}) = \frac1{|N_{a}^{h_T}|}
      \underset{h\in N_{a}^{h_T}}\sum q(h)$ \COMMENT{2in}{=$0$ in case of $N_{a}^{h_T}$\!=$0$}{0.2in}
  \ENDFOR
  \STATE $\widehat{a}_T = \underset{a\in A}\argmax(Q(h_T,{a}))$
  \COMMENT{2in}{compute action with highest value}{0.2in}
  \STATE $\widetilde{a}_T = \underset{a\in A}\argmax\left(\frac1{|N_{a}^{h_T}|}\underset{h\in N_a^{h_T}}\sum\mu(h,h_T)\right)$
  \COMMENT{2.5in}{compute best exploration action $\widetilde{a}_T$
    }{0.1in}
  \IF{randR(0,1) $< \epsilon$}
    \STATE perform $\widetilde{a}_T$
    \COMMENT{2in}{select the exploratory action}{0.2in}
  \ELSE
    \STATE perform $\widehat{a}_T$
    \COMMENT{2in}{select the greedy action}{0.2in}
  \ENDIF
  \FOR {$i = 1$ to $n$}
    \STATE $t =$randZ$(1,T-1)$
    \STATE $q(h_t)\Leftarrow (1-\beta)q(h_t) + \beta(R_t +
                        \gamma{\underset{a}\max Q(h_{t+1},a)})$
  \COMMENT{1.5in}{update the q-values}{0.05in}
  \ENDFOR
  \STATE $q(h_T)=Q(h_T,\widehat{a}_T)$
  \COMMENT{3in}{initialize q-values with a nearest-neighbor estimate}{0.1in}

\ENDLOOP
\end{algorithmic}
\end{algorithm}

\paradot{Piecewise Continuous NSM}
The distance measure used in NSM (equation~\ref{eq:nsm-metric})
was designed for discrete state spaces.  In the continuous
perceptual space where our robot must learn, this metric is
inadequate since most likely all the triples $(a_{t}, o_{t},
R_{t})$ will be different from each other and $\mu(h_{t},
h_{t^{\prime}})$ will always equal 1.  Therefore, to accommodate
continuous states, we replace equation~\ref{eq:nsm-metric} with
the following discounted metric:
\beq\label{eq:discounted-metric}
 \mu(h_{t},h_{t^{\prime}}) = \sum_{\tau=0}^{min(t, t^{\prime})} \lambda^\tau
 ||o_{t-\tau}-o_{t^{\prime}-\tau}||_2,
\eeq
where $\lambda\in [0,1]$. This metric takes an exponentially
discounted average of the Euclidean distance between observation
sequences.  Note that, unlike equation~\ref{eq:nsm-metric}, this
metric ignores actions and rewards.  The distance between action
sequences is not considered because there is no elegant way to
combine discrete actions with continuous observations, and because
our primary concern from a robotics perspective is to provide a
metric that allows the robot to localize itself based on
observations.  Reward values are also excluded to enable the robot
to continue using the metric to select actions even after the
reinforcement signal is no longer available (i.e. after some
initial training period).

Algorithm~\ref{alg:tra} presents PC-NSM in pseudocode.  The
functions randZ(a,b) and randR(c,d) produce a uniformly
distributed random number in $[a,b]\in\mathbb{Z}$ and
$[c,d]\in\mathbb{R}$ respectively, and $\epsilon \in [0,1]$
determines the greediness of the policy.  The algorithm differs
most importantly from NSM in using the discounted metric (line 8),
and in the way exploratory actions in the $\epsilon$-greedy policy
are chosen (line 12).  The exploratory action is the action whose
neighborhood has the highest average distance from the current
observation-state, i.e. the action about which there is the least
information.  This policy induces what has been called {\em
balanced wandering}~\cite{kearns:nearoptimal98}.

\paradot{Endogenous updates}
If the $q$-values are only updated during interaction with the
real environment, learning can be very slow since updates will
occur at the robot's control frequency (i.e.\ the rate at which
the agent takes actions).  One way to more fully exploit the
information gathered from the environment is to perform updates on
the stored history between normal updates. We refer to these
updates as {\em endogenous} because they  originate within the
learning agent,  unlike normal, {\em exogenous} updates which are
triggered by ``real'' events outside the agent.

During learning, the agent selects random times $t<T$, and updates
the $q$-value of $h_t$ according to equation~\ref{eq:update} where
the maximum $Q$-value of the next state $h_{t+1}$ is computed
using equation~\ref{eq:qval} (see lines 18--21 in
Algorithm~\ref{alg:tra}). This approach  is similar to the Dyna
architecture~\cite{Sutton:90dynaold} in that the history acts as a
kind of model, but, unlike Dyna,  the model does not generate new
experiences, rather it re-updates those already in the history in
a manner similar to experience replay~\cite{lin:ml92}.

\section{Experiments in Robot Navigation}\label{sec:exp}

We demonstrate PC-NSM on a mobile robot task where a CSEM Robotics
Smartease\texttrademark\\ robot must use video input to  identify and
navigate to a target object while avoiding obstacles and walls.
Because the camera provides only   a partial view of the environment,
this task requires the robot to use its history of  observations to
remember both where it has been,  and where it last saw the target if
the target moves out of view.

\begin{figure*}[htb]
\includegraphics[width=0.49\textwidth]{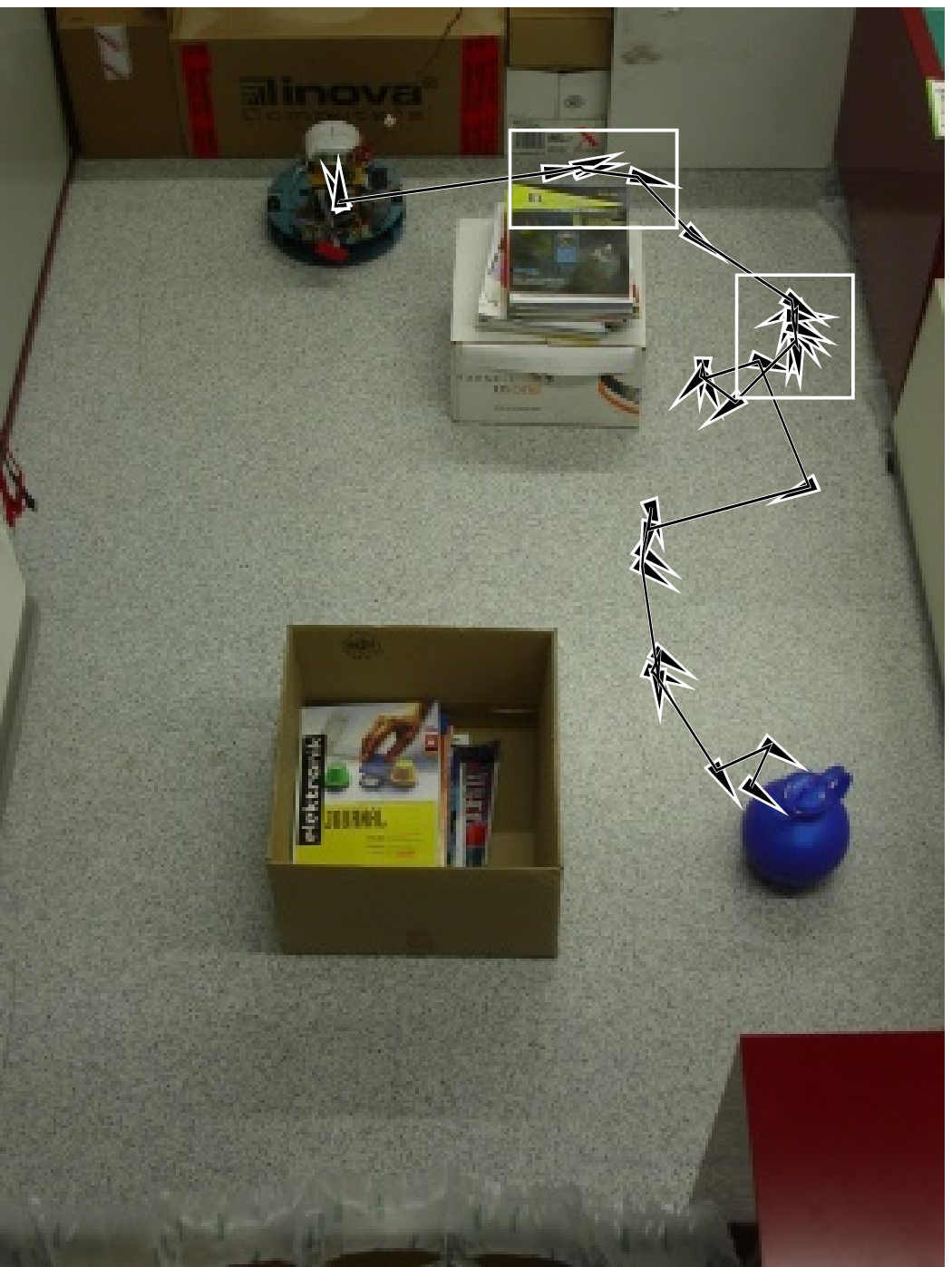}
\includegraphics[width=0.49\textwidth]{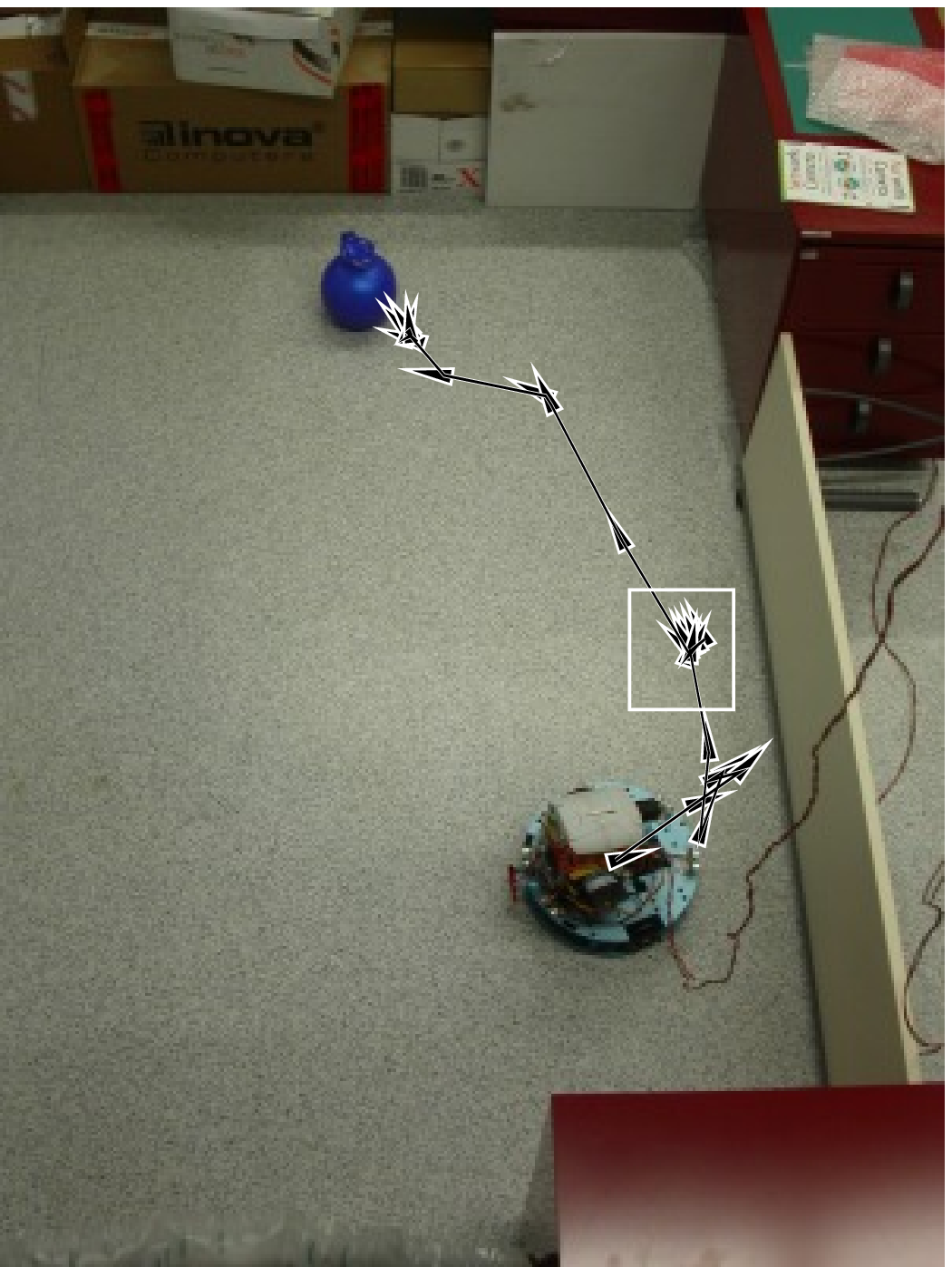}
\vspace{-3mm}
(a) \hspace{2.25in} (b)
\caption{\label{fig:scenario}{\bf Robot arena with learned
trajectories.}
The picture shows two typical learning scenarios with the robot,
obstacle, and target (blue plastic teapot)  in their initial
locations.   The robot must learn to find the target using the
limited visual field of its video camera, and move to it while
avoiding obstacles. (a) The robot starts the trial facing away
from the target before turning to its right, and navigating around
the obstacle to the goal. Along the way it encounters two
difficulties (shown by the rectangular highlights) caused first by
going through the narrow passage and then by gap in the right wall
where the front sonar returns a strong signal compared to the
immediately surrounding wall because it is being specularly
reflected. (b) The robot in an unoccluded arena after learning in
trial (a).  Here the robot drives almost directly to the target
after turning to see it, but, again, is held up by the dark wall
gap momentarily.}
\end{figure*}

\paradot{Experimental Setup}\label{sec:UAIR}
The experiments were  conducted in the 3x4 meter walled arena shown in
figure~\ref{fig:scenario}.  The robot is equipped with two ultrasound
distance sensors (one facing forward, one backward), and a vision
system based on the Axis 2100 network camera that is mounted on top of
the robot's 28cm diameter cylindrical chassis.

\begin{figure}[t]
\begin{center}
\includegraphics[width=0.48\textwidth]{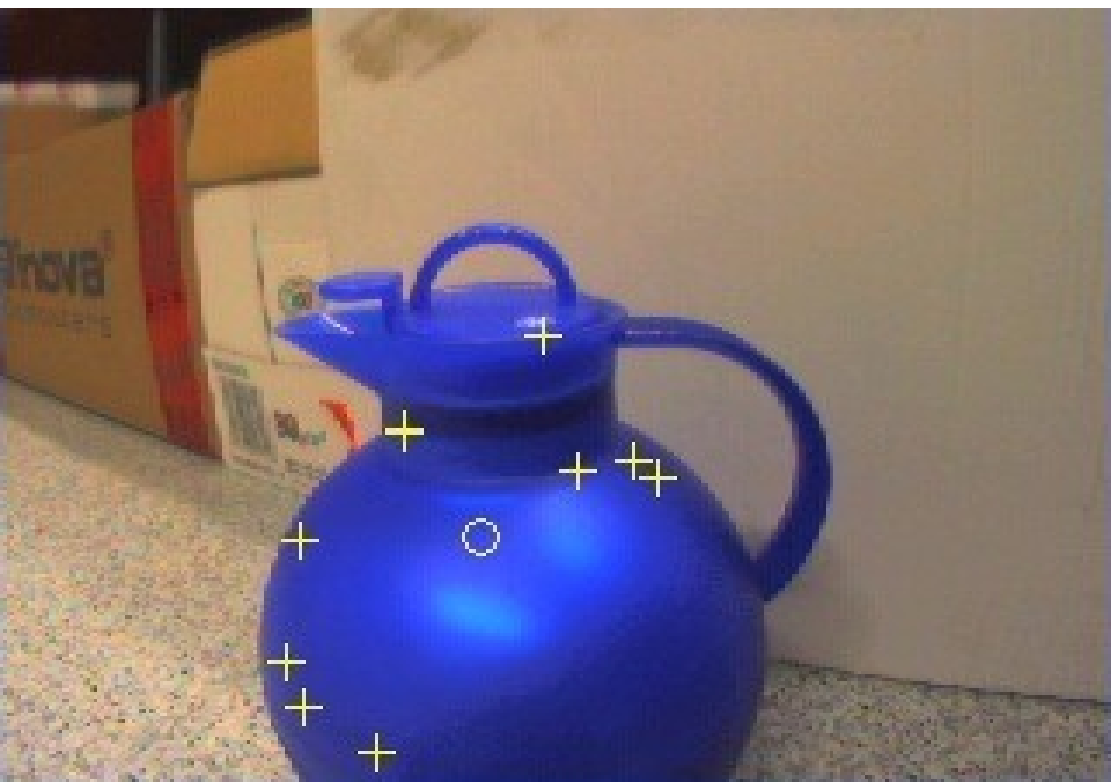}
\includegraphics[width=0.485\textwidth]{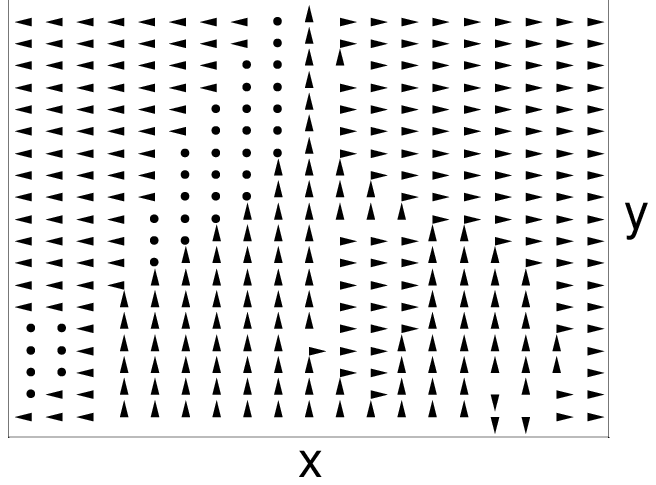}
\includegraphics[width=0.48\textwidth]{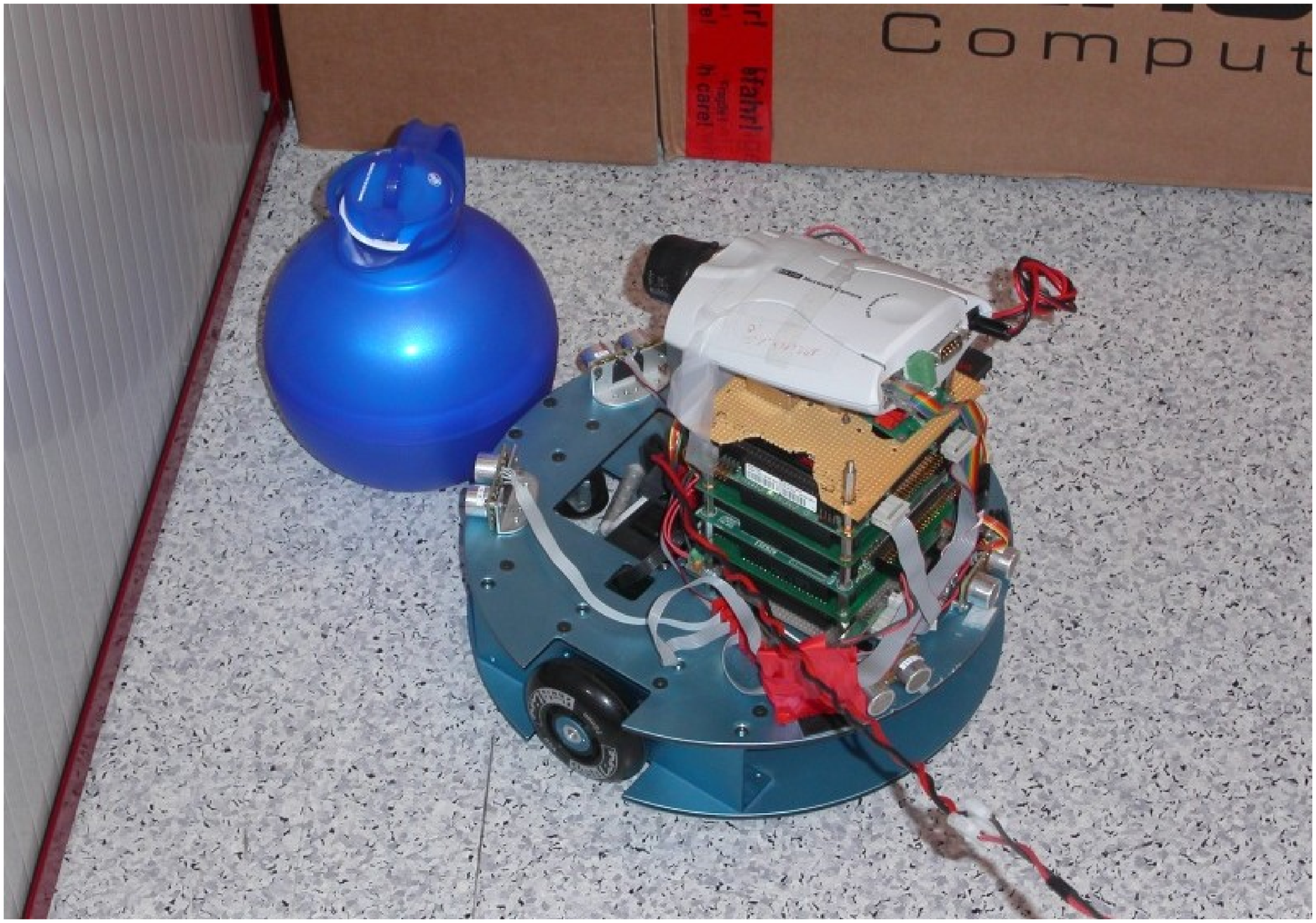}
\includegraphics[width=0.485\textwidth]{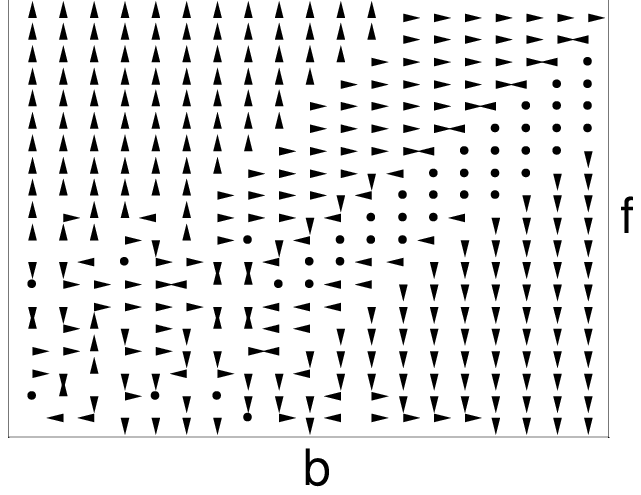}
\end{center}
\vspace*{-5mm} \caption{\label{fig:policy}{\bf Learned Control
Policy.}  Each row shows a different situation in the environment
along with its corresponding learned policy.  In the top row the
robot is positioned directly in front of the target object. The
crosses in the camera image mark detected pixels of the target
color, and the circle indicates the assumed direction towards the
target.  The policy for this situation is shown in terms of the
visual coordinates, i.e. only the $x$-$y$ camera view coordinates
of the high dimensional policy are shown.  Each point in the
policy graph indicates, with an arrow, the direction the robot
should move if the circle, shown in the image is at that point in
the visual field (left arrow means move left, right=right,
up=forward, down=backwards, and no arrow=stand still.  For
instance, in this case, the robot should move forward because the
circle lies in a part of the policy with an up arrow.  In the
bottom row the robot is almost touching the target. Here the
policy is shown in terms of the subspace spanned by the two
ultrasound distance sensors found at the fore and aft of the
robot.  The $b$-axis is the distance from the robot to the nearest
obstacle in front, the $f$-axis behind. When the robot is with its
back to an obstacle, and the way forward is clear (upper left
corner of policy graph), it tends to go forward. When the way
forward is obstructed, but there is nothing behind the robot
(lower right corner), the robot tends to turn or move backward.  }
\end{figure}

Learning was conducted in a series of trials where the  robot,
obstacle(s), and target (blue teapot) were placed at random locations
in the arena.   At the beginning of each trial, the robot takes a
sensor reading and sends, via wireless, the camera  image to a {\em
vision computer}, and the sonar readings to a {\em learning computer}.
The vision computer extracts the $x$-$y$ coordinates of the target in
the visual field by calculating the centroid of $c_p$ pixels
of the target color (see figure\ref{fig:policy}),
and passes them on to the learning computer, along
with a predicate $p$ indicating whether the target is visible. If $p$
is false, $x$=$y$=0.  The learning computer merges $x,y$, and $p$ with
the forward and backward sonar readings, $f$ and $b$, to form the {\bf
inputs} to PC-NSM: an observation vector $o=(x,y,p,f,b)$, where $x$
and $y$ are normalized to $[-1,1]$, and $f$ and $b$ are normalized to
$[0,1]$.

PC-NSM then selects one of 8 {\bf actions}: turn left or right by
either $22.5^\circ$ or $45^\circ$, and  move forward or backward
either 5cm or 15cm (approximately). This action set was chosen to
allow the algorithm to adapt to the scale of
environment~\cite{Schoknecht:03-learning}.  The selected action is
sent to the robot, the robot executes  the action, and the cycle
repeats.  When the robot reaches the goal, the goal is moved to a new
location, and a new trial begins.

The entire interval from sensory reading to action execution is 2.5
seconds, primarily due to camera and network delays.  To accommodate
this relatively low control frequency, the maximum velocity of the
robot  is limited to 10 cm/s.  During the dead time between actions,
the learning computer conducts as many endogenous  updates as time
permits.

\paradot{PC-NSM parameters}
PC-NSM uses an $\epsilon$-greedy policy (Algorithm~\ref{alg:tra},
line 13), with $\epsilon$ set to 0.3.  This means that 30\% of the
time the robot selects an exploratory action.  The appropriate
number of nearest neighbors, $k$, used to select actions, depends
upon the noisiness of the environment.  The lower the noise, the
smaller the $k$ that can be chosen.  For the amount of noise in
our sensors, we found that learning was fastest for $k=3$.

A common practice in toy reinforcement learning tasks such as
discrete mazes is to use minimal reinforcement so that the agent
is rewarded only when it reaches the goal.  While such a
formulation is useful to test algorithms in simulation, for real
robots, this sparse, delayed reward forestalls learning as the
agent can wander for long periods of time without reward, until
finally happening upon the goal by accident.

Often there is specific domain knowledge that can incorporated
into the reward function to provide intermediate reward that
facilitates learning in robotic domains where exploration is
costly~\cite{mataric:icml94}. The reward function we use is the
sum of two components, one is obstacle-related,
$R_{\mbox{\footnotesize obstacle}}$, and the other is
target-related ,$R_{\mbox{\footnotesize target}}$:
\beq\label{eq:reward}
  R \;=\; \underbrace{-20/\max(0.01,\min(f, b))}_{R_{\mbox{\tiny
        obstacle}}} + \underbrace{p\cdot(500 - 50 |x| - 250 y +
        c_p)}_{R_{\mbox{\tiny target}}}
\eeq
$R_{\mbox{\footnotesize target}}$ is largest when the robot is
near to the goal and is looking directly towards it, smaller when
the target is visible in the middle of the field of view, even
smaller when the target is visible, but not in the center, and
reaches its minimum when the target is not visible at all.
$R_{\mbox{\footnotesize obstacle}}$ is negative when the robot is
too close to some obstacle, except when the obstacle is the target
itself, visible by the robot.  It is important to note that the
coefficients in equation~\ref{eq:reward} are specific to the robot
and not the environment.  They represent a one-time calibration of
PC-NSM to the robot hardware being used.

\paradot{Results}\label{sec:ER}
After taking between 1500 and 3000 actions the robot learns to
avoid walls, reduce speed when approaching walls, look around for
the goal, and go to the goal whenever it sees it. This is much
faster compared to neural network based learners,
e.g.~\cite{iida:arob03}, where 4000 episodes were required
(resulting in more than 100'000 actions)  to solve a simpler task
in which the target was always within the perceptual field of the
robot.  Neither do we need a virtual model environment and manual
quantization of the state space like in~\cite{minato}.  To our
knowledge, our results are the fastest in terms of learning speed
and use least quantization effort compared to all other methods to
date, though  we were unable to compare results directly on the
hardware used by these competing approaches.

In the beginning of learning, corners pose serious difficulty
causing the robot to get stuck and receive negative reinforcement
for being too close to a wall.  When the robot accidentally turns
towards the target, it will quickly lose track of it.  As learning
progresses, the robot is able to recover (usually within one
action) when an exploratory action causes it to turn away and
loose sight of the target. The discounted metric allows the robot
to use its history of real-valued observation states to remember
that it had just seen the target in the recent past.
Figure~\ref{fig:policy} shows the learned policy for this task.

\begin{figure}[t]
\begin{center}
\includegraphics[width=\textwidth]{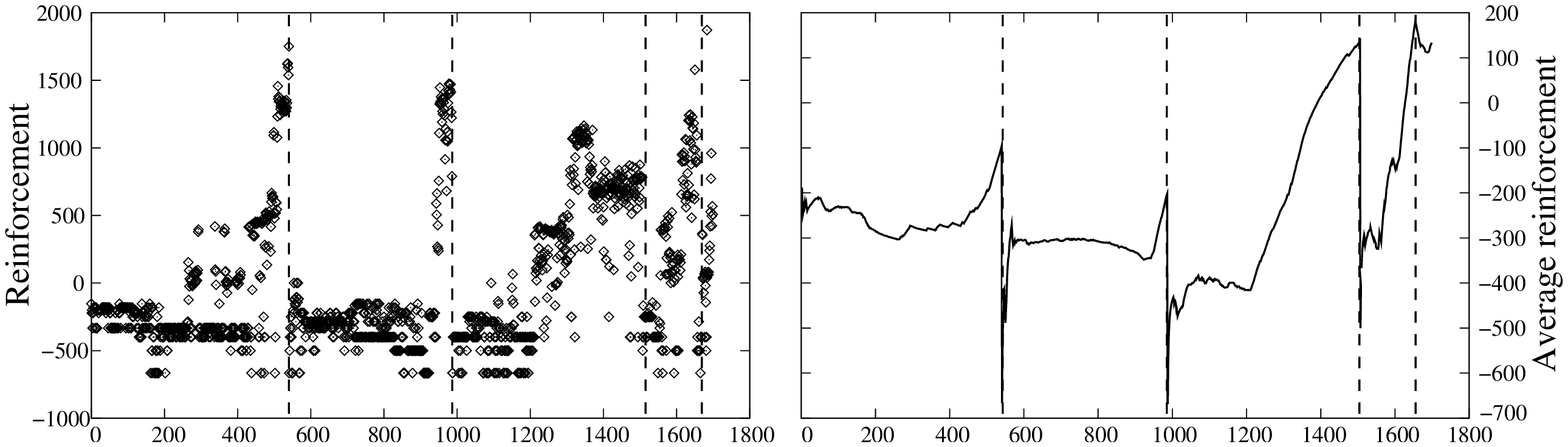}
\end{center}
\vspace{-3mm}
(a) \hspace{2.25in} (b)
\vspace{-2mm}
\caption{\label{fig:reward}{\bf PC-NSM learning performance}.
(a) The plot shows the reward the robot receives at each time-step
during learning.
(b) The plot shows the reward at each time-step averaged over all previous
time-steps within the same trial.
The dashed lines indicate the beginning of a new trial where the target
is moved to a new location.
}
\end{figure}

Since the robot state space is perception-based (not $x$-$y$
coordinates on the floor as is the case in RL textbook examples),
changing the position of the obstacles or target does not impede
robot performance. Figure~\ref{fig:reward} shows learning in terms
of immediate and average reward for a typical sequence of trials
lasting a total of approximately 70 minutes. The dashed vertical
lines in the two graphs indicate the beginning of a new trial. As
learning progresses the robot is able to generalize from past
experience and more quickly find the goal. After the first two
trials, the robot starts to accumulate reward more rapidly in the
third, after which the fourth trial is completed with very little
deliberation. Figure~\ref{fig:scenario} illustrates two such
successful trials.

\section{Discussion}\label{sec:discussion}

We have developed a instance-based algorithm for mobile robot
learning and successfully implemented it on an actual
vision-controlled robot. The use of a metric state space allows
our algorithm to work under weaker requirements and be more
data-efficient compared to previous work in continuous
reinforcement
learning~\cite{doya00reinforcement,Santamaria97experiments,
smart00practical}. Using a metric instead of a discrete grid is a
considerable relaxation of the programmer's task, since it
obviates the need to guess  the correct scale for all the regions
of the state space in advance. The algorithm explores the
environment and learns directly on a mobile robot without using a
hand-made computer model as an intermediate step, works in
piecewise continuous perceptual spaces, and copes with partial
observability.

The metric used in this paper worked well in our experiments, but
a more powerful approach would be to allow the algorithm to select
the appropriate metric for a given environment and task
automatically. To choose between metrics, a criterion should be
defined that determines which of a set of {\it a priori}
equiprobable metrics $\{\mu_1,\dots,\mu_n\}$ fits the given
history of experimentation better. A useful criterion could be,
for example, a generalization of the criteria used in the
McCallum's U-Tree algorithm~\cite{McCallum:96} to decide whether a
state should be split.

The current algorithm uses discrete actions so that there is a
convenient way to group observation states. If the action space
were continuous, the algorithm lacks a natural way to generalize
between actions. A metric on the action space $\mu_a$ could be
used within the observation-based neighborhood delimited by the
current metric $\mu$.  The agent could then randomly sample
possible actions at the query point $h_T$ and obtain Q-values for
each sampled action by computing the $\mu_a$-nearest neighbors
within the $\mu$-neighborhood. Future work will explore this
avenue.

\paradot{Acknowledgments}
This work is partially supported by CSEM Robotics Alpnach.


\end{document}